\documentclass[english]{article}
\usepackage[T1]{fontenc}
\usepackage[latin1]{inputenc}
\usepackage{babel}
\usepackage{amsmath}
\usepackage{amsfonts}
\usepackage{mathenv}
\usepackage{graphicx}
\usepackage{subfigure}
\usepackage{hyperref}
\usepackage{authblk}
\numberwithin{equation}{section}

\newtheorem{lemma}{Lemma}[section]

\newtheorem{definition}[lemma]{Definition}

\title{Toward better feature weighting algorithms: \\ a focus on Relief}
\author{Gabriel Prat Masramon}
\author{Llu\'is A. Belanche Mu{\~n}oz}
\affil{Departament de Llenguatges i Sistemes Inform\`{a}tics (LSI), Universitat Polit\`{e}cnica de Catalunya (UPC)}

\date{15\textsuperscript{th} June 2005}

\begin{document}

\maketitle 
\begin{abstract}
Feature weighting algorithms try to solve a problem of great importance nowadays
in machine learning: The search of a relevance measure for the features of a given
domain. This relevance is primarily used for feature selection as feature weighting
can be seen as a generalization of it, but it is also useful to better understand a 
problem's domain or to guide an inductor in its learning process. Relief family of
algorithms are proven to be very effective in this task. Some other feature weighting
methods are reviewed in order to give some context and then the different existing
extensions to the original algorithm are explained. 

One of Relief's known issues is the performance degradation of its estimates when 
redundant features are present. A novel theoretical definition of redundancy level 
is given in order to guide the work towards an extension of the algorithm that is 
more robust against redundancy. A new extension is presented that aims for improving 
the algorithms performance. Some experiments were driven to test this new extension 
against the existing ones with a set of artificial and real datasets and denoted that
in certain cases it improves the weight's estimation accuracy.
 
\end{abstract}
\section{Overview}
Feature selection is undoubtedly one of the most important problems in machine 
learning, pattern recognition and information retrieval, among others. A feature 
selection algorithm is a computational solution that is motivated by a certain 
definition of relevance.
However, the relevance of a feature may have several definitions depending on
the objective that is looked after.

The generic purpose pursued is the improvement of the inductive
learner, either in terms of learning speed, generalization
capacity or simplicity of the representation. It is then
possible to understand better the obtained results, diminish the
volume of storage, reduce noise generated by irrelevant or
redundant features and eliminate useless knowledge.

On the other hand, feature weighting algorithms try to estimate relevance (in the form 
of weights to the features) rather than binarily deciding whether a feature is either 
relevant or not. This is a much harder problem, but also a more flexible framework 
from an inductive learning perspective. This kind of algorithms are confronted with
the down-weighting of irrelevant features, the up-weighting of relevant ones and the
problem of relevance assignment when redundancy is an issue. 

In this work we review Relief, one of the most popular feature weighting algorithms. 
After a state-of-the-art in section \ref{sec:state_of_the_art} focused on feature 
weighting methods in general, in section we describe the algorithm and its more 
important extensions. We are primarily interested in coping
with redundancy, and studying to what extent can the Relief algorithm be modified in 
order to better its treatment of redundancy, which is one of its known weaknesses. 
In this vein, section \ref{sec:new} points out a novel and general (though 
computationally infeasible) definition of redundancy level and try to relate it to 
the actual Relief performance. Next, we develop a "double" or feedback extension  
of the algorithm that takes its own estimations into account in order to improve 
general performance. We also complement this matter with a set of experiments in 
section \ref{sec:empirical}. The work concludes with some open questions and 
clear avenues of continuation of the material herein presented.

\section{State of the art}\label{sec:state_of_the_art}
\subsection{Introduction}
In the last few years feature selection has become a more and more common topic of research. This 
popularity increase is probably due to the growth of the problem domains' number of features. No 
more than ten years ago few problems treated domains with more than 50 features. Nowadays most 
papers deal with domains with hundreds and even tens of thousands of features. New techniques have 
to be developed to address this kind of problems with many irrelevant and redundant features and  
comparatively few instances to learn from. One example of these new domains is web page categorization,
a domain currently of much interest for internet search engines where thousands of terms can be found 
in a document. Another example can be appearance-based image classification methods which may use
every pixel in the image. Classification problems with thousands of features are very common in 
medicine and biology; e.g. molecule classification, gene selection or medical diagnostics. In medical
problems we typically have less than a hundred patients and for each patient we can have thousands
of features evaluated.

Feature selection can help us solving a classification problem with these characteristics for many
reasons. Firstly it may make the task of data visualization and understanding easier by eliminating 
irrelevant features which can mislead the interpretation of the data. It can also reduce the cost of
the measurements as we can avoid measuring irrelevant features; this is especially important in domains
where some features are very expensive to obtain, e.g., require a special medical test. In addition, 
a big benefit of feature selection is defying the curse of dimensionality to help the induction of 
good classifiers from the data. When many {\it unuseful}, i.e. irrelevant or redundant,  features are 
present in training data, classifiers may find false regularities in the input features and learn from 
that instead of learning from the features that really determine the instance class (also valid when 
predicting the instance target value in the case of regression). 
 
\begin{figure}[htbp]
\centering
\fbox{\includegraphics{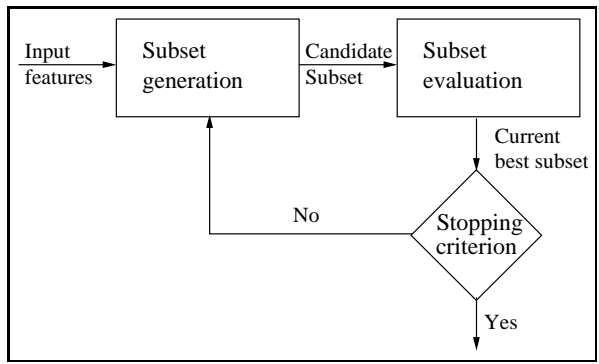}}
\caption{Feature selection framework}
\label{fig:feat_sel_framework}
\end{figure}

There are two main approaches to feature selection: filter methods and wrapper methods. Both methods
can be included in the framework shown on Fig. \ref{fig:feat_sel_framework}. The main difference 
between them is the use of a classifier for the estimation of a feature {\it usefulness}.
The two families of methods only differ in the way they evaluate the candidate sets of features. While
the former methods use a problem independent criterion, the latter use the performance of the final 
classifier to evaluate the quality of a feature subset. The basic idea of the filter methods is to 
select the features according to some prior knowledge of the data. For example, to select the features
based on the conditional probability that a given instance is a member of a certain class given the
value of its features. Another criterion commonly used by filter methods is the correlation of a
feature with the class, i.e. selecting features with high correlation. More detailed criteria is given
in section \ref{sec:feat_weight}, where also more criteria are described. In contrast, wrapper methods 
suggest a set of features that are given to a classifier which uses them to classify some training data 
and returns the performance of the classification which is the acceptance criterion of the feature set.

Now we have explained two approaches of feature subset evaluation, but is clear to see that if we had to
test all possible subsets, using either of the methods, of features we would have a combinatorial explosion.  
If our initial set of features is $\mathcal{F}$ and $|\mathcal{F}|=n$, the number of evaluations we would
have to do would be equal to the cardinality of the power set of $\mathcal{F}$: 
$|\mathcal{P}(\mathcal{F})| = 2^n$. For this reason diverse techniques have been developed to reduce the
computational complexity of this problem. 

A different technique of determining feature {\it usefulness} apart from feature selection is a technique
called feature weighting (or feature ranking). It consists of assigning a numeric value to each feature 
so as to indicate the feature's {\it usefulness}. Feature weighting can help solving the problem of 
feature selection. One possible approach to feature selection using feature weighting could be to first
assign weights to features and then choose features according to their weights. This can be done either 
by having a rule to binarize the weights, e.g. select all the features with weight greater than zero, or
by means of a weight guided feature subset evaluation, e.g. evaluating the subsets containing the features
with greatest weight values. In fact, feature weighting could be seen as a generalization of feature 
selection, i.e. feature selection would be a specific kind of feature weighting where the weights assigned 
to features are binary.

In following sections we will explore various methods of existing feature weighting algorithms than and will 
discuss their properties to later have some starting point to describe and analyze the algorithm in the focus 
of this paper: Relief.

\subsection{Feature weighting}\label{sec:feat_weight}
This section will review some of the most used feature weighting algorithms. Although the section is focused
on feature weighting, most of the methods described below can also be used for feature selection.

On following subsections $\mathcal{I}$ and $\mathcal{F}$ represent the sets of instances
and features respectively. $I$, $I_1$ or $I_i$ represent instances from $\mathcal{I}$. $X$, $X_i$ or
$Y$ are sets of possible feature values from a feature in $\mathcal{F}$. $C$ represents the set of possible 
class values.  And their lower case versions represent single value in its correspondent upper case set, 
e.g. we will use $c \in C$ and $x \in X$. We also will use a short notation to express probabilities, e.g. 
will write $p(x)$ to represent the probability for feature $X$ to have value $x$ or $p(c|x)$ to express
the conditional probability of the class to have value $c$ knowing that the feature $X$ has value $x$.

\subsubsection*{Conditional Probabilities based methods}
The first group of methods we will look at are the ones based on conditional probabilities of class given
a feature value. Two simple methods using this idea were introduced in \cite{ACM:Creecy1992}: per-category 
feature importance and cross-category feature importance (or, in short, PCF and CCF). One important 
limitation is that they can only deal with binary features, so numerical features must be discretized
and symbolic features converted to a group of binary features. The weights assigned to features in the
case of PCF depends on the class of the feature as seen in Eq. \ref{eq:pcf} 
\begin{equation}
w_{PCF}(X,c)=P(c|x)\text{, where $x$ would be the {\it positive} feature value}
\label{eq:pcf}
\end{equation}
so we have a weight for each feature and class. CCF relies on the same idea but instead of having one 
weight for each feature and class it have
only a weight per feature. It does so by averaging the weights across classes. In fact, as it shows Eq.
\ref{eq:ccf}, it uses the summation of squares of conditional probabilities.
\begin{equation}
w_{CCF}(X)=\sum_{c \in C}{P(c|x)^2}\text{, where $x$ would be the {\it positive} feature value}
\label{eq:ccf}
\end{equation}
Later on \cite{mohri94optimal} showed that PCF is too sensitive to class proportions and tends to answer
the most frequent class when using it for classifying. 

A more sophisticated approach that also makes use of conditional probabilities is the one used by the value 
difference method (VDM) introduced by \cite{DBLP:journals/cacm/StanfillW86}. This time no binarization of 
features is required, although numeric features still have to be discretized in order to calculate 
conditional probabilities as shown in Eq. \ref{eq:vdm}. In addition this method does not assign weights to
each feature but to each value of each feature.
\begin{equation}
w_{VDM}(X,x) = \sqrt{\sum_{c \in C}{\left({\frac{P(x|c)}{p(x)}}\right)^2}}
\label{eq:vdm}
\end{equation}
This weighting scheme was originally used to calculate distances between features.

Finally we have Gini-index gain \cite{DBLP:books/wa/BreimanFOS84} in Eq. \ref{eq:gini} which can 
be interpreted as the expected error rate
\begin{equation}
GG(X) = \sum_{x \in X}{P(x) \sum_{c \in C} {P(c|x)^2}} - \sum_{c \in C} {P(c)^2} 
\label{eq:gini}
\end{equation}
and is proven to be biased towards multiple valued features. In further sections we will see that this
particular measure has some relation with the Relief algorithm.

\subsubsection*{Information theory based methods}
Not all the feature weighting methods are based on conditional probabilities, though. Now we will describe
some methods based on information theory \cite{Shannon1948,ShannonWeaver1949}.

The first one is just using Shannon's mutual information (MI) between two features $X$ and $Y$
in Eq. \ref{eq:mi}, 
\begin{equation}
MI(X,Y) = H(X) - H(X|Y) = \sum_{x \in X, y \in Y}{p(x,y)log_2\frac{p(x,y)}{p(x)p(y)}}
\label{eq:mi}
\end{equation}
which is defined using entropies and conditional entropies (see Eq. \ref{eq:shannon}),
\begin{align}
\lefteqn \nonumber \text{Entropy: } & & H(X) = -\sum_{x \in X} {P(x)\log _2 P(x)} \\
\label{eq:shannon} \text{Conditional entropy: } & & H(X|Y) = H(X,Y) - H(Y) \\
\nonumber \text{Joint entropy: } & & H(X,Y) = -\sum_{x \in X, y \in Y} {P(x,y)\log _2 P(x,y)}
\end{align}
to weight features. A more informal but maybe more intuitive definition of mutual information is that
MI measures the information of $X$ that is also in $Y$. If the features are independent no information
is shared so mutual information is zero. In the other end we have that one feature is an exact copy of
the other, all the information it contains is also shared by the other so the mutual information is the
same as the information conveyed by one of them, namely its entropy. A very popular feature weighting
method uses the idea of mutual information. It was proposed by \cite{PRNN:Hunt+Marin+Stone:1966}
and it is used in \cite{DBLP:journals/ml/Quinlan86} when splitting nodes in top down indutcion of 
decision trees (TDIDT) best known as ID3. The term information gain (IG) in Eq. \ref{eq:inf_gain}
is used there. Its intuitive interpretation would be: The more an feature reduces class entropy when 
knowing its value, the more its weight. This is just another way to say: The more information is shared 
between an feature and the class, the more its weight. So if we have a set of classes $C$ we can define
IG for the class knowing the value of a feature $X$ as shown in Eq. \ref{eq:inf_gain}
\begin{equation}
IG(C|X) = MI(C,X).
\label{eq:inf_gain}
\end{equation}

Later on, similar methods were introduced to reduce the bias of IG towards features with large number 
of values. The extreme case is using an feature with an ID code. It is clear to see that knowing the 
ID code we can precisely know the class of any instance in our training set. The problem is that we can
say nothing about a new instance which will have another unknown ID code. One of these methods is gain 
ratio (GR) in Eq. \ref{eq:gain_ratio} used by C4.5 decision tree induction algorithm \cite{ACM:Quinlan1993} 
which normalizes IG by the amount of information needed to predict an features value (the entropy of 
the feature). But there are also various other proposals, among them there are entropy distance 
\cite{MacKay:itp} in Eq. \ref{eq:entropy_distance} and the Mántaras distance between the class and the
 feature in Eq. \ref{eq:mantaras} which was proved to be unbiased towards multiple-valued features.
\begin{align}
GR(C|X) & = \frac{IG(C|X)}{H(X)} \label{eq:gain_ratio} \\
D_H(C,X) & =  H(C,X) - MI(C,X) \label{eq:entropy_distance} \\
D_M(C,X) & = \frac{H(X|C)+H(C|X)}{H(C,X)} = 2-\frac{H(X)+H(C)}{H(C,X)}\label{eq:mantaras}
\end{align}

\subsubsection*{Distribution distance based methods}
Another way to find dependencies between a feature and the class is to measure differences between their
distributions. Perhaps the simplest way to do so is to compute the difference between the joint and the
product distributions as shown in Eq. \ref{eq:diff_dist}
\begin{equation}
\operatorname{Diff}(C,X) = \sum_{c \in C, x \in X} {|P(c,x) - P(x)P(c)|}
\label{eq:diff_dist} 
\end{equation} 
and this distance can be directly used as the features weight. Large differences between the joint 
and the product distributions indicate large dependency of the class on the feature, so the feature
should be given a large weight. This can easily be applied to continuous features changing the sum for
an integration. It can also easily be rescaled to the [0,1] interval as it has an upper bound of 
$1-\sum_{x\in X}{P(x)^2}$.

More distance functions can be used here. An interesting one is the Kullback-Leibler divergence which 
is not a distance in fact as it is not symmetric (i.e., $D_{KL}(X||Y)\neq D_{KL}(Y||X)$). The application
on feature weighting is to have the weigh be equal to the \textit{distance} between the joint and the
product distributions, see Eq. \ref{eq:kullback}.
\begin{equation}
D_{KL}(P(X,C)||P(X)P(C))=\sum_{c \in C, x \in X}{P(c,x)\log \frac{P(c,x)}{P(x)P(c)}} 
\label{eq:kullback}
\end{equation}
Note that this is exactly the same as the mutual information between the feature and the class (see Eq. 
\ref{eq:mi}) so we have $D_{KL}(P(X,C)||P(X)P(C))=MI(X,C)$.

\subsubsection*{Correlation based methods}
Even though this approach to feature weighting is treated last, maybe is one of the simplest as it does
not care about continuous feature discretization or probability density estimations. It is usual in 
statistics to construct contingency tables for pairs of discrete variables to analyze their correlation.
In our case (see Table \ref{taula_contingencia}) we will define a contingency table between the set of 
classes $c_i \in C$ and the values of a feature $x_j \in X$. The inner cells in row $i$ and column $j$ 
of the table contain the number of instances of class $c_i$ that have feature $X=x_j$. The row marginal 
totals will tell the number of instances for the corresponding class and the column marginal totals the 
number of instances with the corresponding value on feature $X$. Finally the sum of either marginal
totals should be the total number of instances $m$.  
% Table: 6 Columns x 5 Rows.
\begin{table}[htbp]
\centering
\begin{tabular}{|l||l|l|l|l||l|}
\hline & $x_1$ & $x_2$ & $\ldots$ & $x_v$ &  Tot. \\
\hline \hline $c_1$ & $N_{11}$ & $N_{12}$ & $\ldots$ & $N_{1v}$ & $N_{1\cdot}$ \\
\hline $\vdots$& \multicolumn{4}{c||}{$\ddots$}& $\vdots$ \\
\hline $c_w$ & $N_{w1}$ & $N_{w2}$ & $\ldots$ & $N_{wv}$ & $N_{w\cdot}$ \\
\hline \hline Tot. & $N_{\cdot 1}$& $N_{\cdot 2}$& $\ldots$& $N_{\cdot v}$ & $m$ \\
\hline
\end{tabular}
\begin{tabular}{ll}
$v$ & No. of values for $X$ \\
$w$ & No. of classes ($C$) \\
$m$ & Total no. of instances \\
$N_{c_i\cdot}$ & Total no. in class $c$ \\
$N_{\cdot x_j}$ & Total no. with $X=x_j$ \\
$N_{c_ix_j}$ & No. with $C=c \wedge X=x_j$
\end{tabular}
\caption{Contingency table of the class vs. the $X$ feature values}
\label{taula_contingencia}
\end{table}
Looking at this table we can define chi-squared weight for feature $X$ as shown on Eq. \ref{eq:chi}
\begin{equation}
X^2(X)=\sum_{x \in X, c \in C}{(N_{cx}-E_{cx})^2/E_{cx}}
\label{eq:chi}
\end{equation}
where $E_{cx}$ is the expected number of instances of class $c$ with value $x$ on feature $X$ calculated
as $N_{c\cdot}N_{\cdot x}/m$. $X^2$ is distributed approximately as a $\chi^2$ with $(v-1)(w-1)$ degrees 
of freedom. We should avoid terms with $E_{cx}=0$ or replace them with a small positive number. We can 
see that in the extreme case that $X$ and $C$ are completely independent  $N_{cx}=E_{cx}$ is expected so
large values of $X^2(X)$ indicate strong dependence between the feature and the class. Note that the 
result of $X^2$ depends not only on the joint probabilities $P(c,x)=N_{cx}/m$ but also depends on the 
number of instances $m$. This dependency on the number of instances seems to make sense with the 
intuition that correlations calculated with small number of instances shall be less accurate.

\subsection{Relief}
One common characteristic of the previously cited methods is that they treat features individually 
assuming conditional independence of features upon the class. In the other hand, Relief takes all
other features in care when evaluating a specific feature. Another interesting characteristic of 
Relief is that it is aware of contextual information being able to detect local correlations of
feature values and their ability to discriminate from an instance of a different class. 

The main idea behind Relief is to assign large weights to features that contribute in separating 
near instances of different class and joining near instances belonging to the same class. The word
"near" in the previous sentence is of crucial importance since we mentioned that one of the main
differences between Relief and the other cited methods is the ability to take local context into 
account. Relief does not reward features that separate (join) instances of different (same) 
classes in general but features that do so for near instances. 

\begin{figure}[htbp]
Input: for each training instance a vector of feature values and the class value

Output: the vector W of estimations of the qualities of features
\begin{enumerate}
\item set all weights $W[A]:=0.0$; 
\item \textbf{for} $i:=1$ \textbf{to} $m$ \textbf{do begin} 
\item \hspace{2em}randomly select an instance $Ri$; 
\item \hspace{2em}find nearest hit $H$ and nearest miss $M$; 
\item \hspace{2em}\textbf{for} $A:=1$ \textbf{to} $a$ \textbf{do}  
\item \hspace{4em}$W[A]:=W[A] - \operatorname{diff}(A,Ri,H)/m + \operatorname{diff}(A,Ri,M)/m$
\item \textbf{end}; 
\end{enumerate}
\caption{Pseudo code of the original Relief algorithm}
\label{fig:relief}
\end{figure}

In Fig. \ref{fig:relief} we can see the original algorithm presented by Kira and Rendell in
\cite{DBLP:conf/aaai/KiraR92}. We maintained the original notation that slightly differs from 
the used above as now features (attributes) are labeled $A$. There we can see that in the aim 
of detecting whether the feature is useful to discriminate near instances it selects two nearest
neighbors of the current instance $R_i$. One from the same class $H$ called the nearest hit and 
one from the different class $M$ (the original Relief algorithm only dealt with two class 
problems) called the nearest miss. With these two nearest neighbors it increases the weight of
the feature if it has the same value for both $R_i$ and $H$ and decreases it otherwise. The opposite 
occurs with the nearest miss, Relief increases the weight of a feature if it has opposite values 
for $R_i$ and $M$ and decreases it otherwise.

One of the central parts of Relief is the difference function $\operatorname{diff}$ which is
also used to compute the distance between instances as shown in Eq. \ref{eq:relief_distance}.
\begin{equation}
\delta(I_1,I_2)=\sum_{i}{\operatorname{diff}(A_i,I_1,I_2)}
\label{eq:relief_distance}
\end{equation}
The original definition of $\operatorname{diff}$ was an heterogeneous distance metric composed
of the \textit{overlap} metric in Eq. \ref{eq:overlap} for nominal features and the 
normalized Euclidean distance in Eq. \ref{eq:euclidean} for linear features, which 
\cite{DBLP:journals/jair/WilsonM97} called HEOM.
\begin{equation}
\operatorname{diff} (A,I_1 ,I_2 ) =\begin{cases}
0&\text{if $\operatorname{value}(A,I_1) = \operatorname{value}(A,I_2)$}\\
1&\text{otherwise}   
\end{cases}
\label{eq:overlap}
\end{equation}
\begin{equation}
\operatorname{diff} (A,I_1 ,I_2 ) = \frac
	{\left| \operatorname{value} (A,I_1) - \operatorname{value}(A,I_2) \right|}
	{\max(A)-\min(A)}
\label{eq:euclidean}
\end{equation}
The difference normalization with $m$ guarantees that the weight range is [-1,1]. In fact the
algorithm tries to approximate a probability difference in Eq. \ref{eq:dif_prob_relief}.
\begin{align}
\nonumber W[A] \approx & P(\text{different value of }A|\text{nearest instance from different class}) - \\
	& P(\text{different value of }A|\text{nearest instance from same class})
\label{eq:dif_prob_relief}
\end{align}
We can see that for a set of instances $\mathcal{I}$ having a set of features $\mathcal{F}$ this 
algorithm has cost $O(m \times |\mathcal{I}| \times |\mathcal{F}|)$ as it has to loop over $m$ instances.
For each instance in the main loop it has to compute its distance from all other instances so we have
$O(m \times |\mathcal{I}|)$ times the complexity of calculating $D_{Relief}$ and we can easily see from 
Eq. \ref{eq:relief_distance} that its complexity is  $O(|\mathcal{F}|)$, so we have our complexity: 
$O(m \times |\mathcal{I}| \times |\mathcal{F}|)$. As $m$ is a user defined parameter we can in some measure
control the cost of Relief algorithm having a tradeoff between accuracy of estimation (for large $m$)
and low complexity of the algorithm (for small $m$). However $m$ can never be greater than $|\mathcal{I}|$.

\subsection{Extensions of Relief}
The first modification proposed to the algorithm is to make it deterministic by changing the outer loop
through $m$ randomly chosen instances for a loop over all instances. This obviously increases the 
algorithms computation cost which becomes $O(|\mathcal{I}|^2 \times |\mathcal{F}|)$ but makes 
experiments with small datasets more reproducible. Kononenko uses this simplified version of the algorithm 
in its paper \cite{DBLP:conf/ecml/Kononenko94} to test his new extensions to the original Relief. This 
version is also used by other authors \cite{DBLP:journals/ai/KohaviJ97} and its given the name \textit{Relieved}
with the final \textit{d} for "deterministic". 

We can find some extensions to the original Relief algorithm proposed in \cite{DBLP:conf/ecml/Kononenko94}
in order to overcome some of its limitations: It couldn't deal with incomplete datasets, it was very sensible
to noisy data and it could only deal with multi-class problems by splitting the problem into series of 
2-class problems.

To able Relief to deal with incomplete datasets, i.e. that contained missing values, a modification of the
$\operatorname{diff}$ function is needed. The new function must be capable of calculating the difference
between a value of a feature and a missing value and between two missing values in addition to the 
calculation of difference between two known values. Kononenko proposed various modifications of this function
in its paper and found one that performed better than the others it was the one in a version of Relief he 
called RELIEF-D (not to be confused with \textit{Releaved} mentioned above). The difference function used
by RELIEF-D can be seen in Eq. \ref{eq:relief_d}.
\begin{equation}
\operatorname{diff}(A,I_1,I_2) =  
\begin{cases}
1 - P(value(A,I_2)|class(I_1)) & \text{ if $I_1$ is missing } \\
1 - \sum\limits_{a \in A}{[P(a|class(I_1)) \times P(a|class(I_2))]} & \text{ if both missing}
\end{cases}
\label{eq:relief_d}
\end{equation}

Now we will focus on giving Relief greater robustness against noise. This robustness can be achieved by
increasing the number of nearest hits and misses to look at. This mitigates the effect of choosing a 
neighbor that would not have been the nearest without the effect of noise. The new algorithm has a new
user defined parameter $k$ that controls the number of nearest neighbors to use. In choosing $k$ there is a 
tradeoff between locality and noise robustness. \cite{DBLP:conf/ecml/Kononenko94} states that 10 is a good 
choice for most purposes.

The last limitation was that the algorithm was only designed for 2-class problems. The straightforward
extension to multi-class problems would be to take as the near miss the nearest neighbor belonging to a
different class. This variant of Relief is the so-called Relief-E by Kononenko. But later on he proposes
another variant which gave better results: This was to take the nearest neighbor (or the $k$ nearest) from
each class and average their contribution so as to keep the contributions of hits and misses symmetric and 
between the interval [0,1]. That gives the Relief-F (ReliefF from now on) algorithm seen in Fig. 
\ref{fig:relieff}.
\begin{figure}[htbp]
Input: for each training instance a vector of feature values and the class value

Output: the vector W of estimations of the qualities of features
\begin{enumerate}
\item set all weights $W[A]:=0.0$; 
\item \textbf{for} $i:=1$ \textbf{to} $m$ \textbf{do begin} 
\item \hspace{2em}randomly select an instance $R_i$; 
\item \hspace{2em}find $k$ nearest hits $H_j$;
\item \hspace{2em}\textbf{for} each class $C \neq class(R_i)$ \textbf{do}
\item \hspace{4em}find $k$ nearest misses $M_j(C)$; 
\item \hspace{2em}\textbf{for} $A:=1$ \textbf{to} $a$ \textbf{do}  
\item \hspace{4em}$W[A]:=W[A] - \sum\limits_{j=1}^k\operatorname{diff}(A,R_i,H_j)/(m\cdot k)+$
\item \hspace{6em}$\sum\limits_{C\neq class(R_i)} \left[\frac{P(C)}{1-P(class(R_i))} \\
						\sum\limits_{j=1}^k\operatorname{diff}(A,R_i,M_j(C))\right]/(m\cdot k)$;
\item \textbf{end}; 
\end{enumerate}
\caption{Pseudo code of the ReliefF algorithm}
\label{fig:relieff}
\end{figure}

The above mentioned relation to impurity functions, in specific with Gini-index gain in Eq. \ref{eq:gini},
can be seen in \cite{DBLP:journals/ml/Robnik-SikonjaK03} when developing the probability difference in
Eq. \ref{eq:dif_prob_relief} in the case that the algorithm uses a large number of nearest neighbors 
(i.e., when the selected instance could be anyone from the set of instances). This version 
of the algorithm is called myopic ReliefF as it loses its context of locality property. Rewriting 
Eq. \ref{eq:dif_prob_relief} by removing the neighboring condition and by applying Bayes' rule, 
we obtain Eq. \ref{eq:dif_prob_relief_short}.
\begin{equation}
W'[A]=\frac{P_{samecl|eqval}P_{eqval}}{P_{samecl}} - \frac{(1-P_{samecl|eqval})P_{eqval}}{1-P_{samecl}}
\label{eq:dif_prob_relief_short}
\end{equation}
For sampling with replacement we obtain we have:
\begin{align*}
P_{eqval} & = \sum\limits_{c \in C}{P(c)^2} \\
P_{samecl|eqval} & = \sum\limits_{x \in X}{\left( \frac{P(x)^2}{\sum_{x \in X}{P(x)^2}} \times 
					\sum\limits_{c \in C}{P(c|x)^2}\right)}
\label{eq:dif_prob_relief_short}
\end{align*}
Now we can rewrite Eq. \ref{eq:dif_prob_relief_short} to obtain the myopic Relief weight estimation:
\begin{equation}
W'[A]=\frac{P_{eqval}\times GG'(X)}{P_{samecl}1-P_{samecl}}
\label{eq:myopic_relief}
\end{equation}
Where $GG'(A)$ is a modified Gini-index gain of attribute $A$ as seen in Eq. \ref{eq:modif_gini}.
\begin{equation}
GG'(X) = \sum_{x \in X}{\left(\frac{P(x)^2}{\sum_{x \in X}{P(x)^2}} \times \sum_{c \in C} {P(c|x)^2}\right)} - \sum_{c \in C} {P(c)^2 } 
\label{eq:modif_gini}
\end{equation}
As we can see the difference in this modified version from its original Gini-index gain described 
above in Eq. \ref{eq:gini} is that Gini-index gain used a factor:
\begin{equation*} 
\frac{P(x)}{\sum_{x \in X}{P(x)}} = P(x)
\end{equation*} 
while myopic ReliefF uses:
\begin{equation*} 
\frac{P(x)^2}{\sum_{x \in X}{P(x)^2}}
\end{equation*}

So we can see how this myopic ReliefF in Eq. \ref{eq:myopic_relief} holds some kind of normalization
for multi-valued attributes when using the factor $P_{eqval}$. This solves the bias of impurity functions
towards attributes with multiple values. Anther improvement compared with Gini-index is that Gini-index
gain values decrease when the number of classes increase. The denominator of Eq. \ref{eq:myopic_relief} 
avoids this strange behavior.

\section{New apportations}\label{sec:new}
\subsection{Redundancy analysis}
To begin with the redundancy analysis of Relief, we first of all have to define 
exactly the meaning of redundancy. In general the definitions of redundancy we
find in the literature are based on feature correlation, i.e. two features are 
redundant if their values are correlated. One interesting particular case is when
one feature is an exact copy of another so their values are completely correlated,
one feature is obviously redundant. But in reality a feature may not be completely
correlated with another feature but may be (partially) correlated with a set of
features. In such case it's not straightforward to determine redundancy. We can take
as an example the features shown in Table~\ref{taula_red}. The feature $f_r$ is 
intuitively redundant with the set $\{f_1,f_2\}$ but is not correlated with any of them,
so it would not be redundant according to the correlation based definition of redundancy.
\begin{table}[htbp]
\centerline{
\begin{tabular}{|ccc||c|}
	\hline
$f_1$    &   $f_2$  & $f_r$ & $C$    \\
	\hline \hline
0 & 0 & 1 & 0 \\
0 & 1 & 1 & 0 \\
1 & 0 & 1 & 0 \\
1 & 1 & 0 & 1 \\
	\hline
\end{tabular}}
\caption{Two relevant and one redundant features: $C=f_1 \wedge f_2$ and $f_r=\overline{f_1 \wedge f_2}$}
\label{taula_red}
\end{table}
So we have to find a better definition for feature redundancy that enables us to 
identify not only pairs of redundant features but features redundant with any set
of other features. Before giving the formal definition of redundancy let's introduce
some previous definitions:
\begin{definition}
Let $\mathbf{U} = \{\alpha, \beta, \ldots\}$ be a set of discrete variables in a problem domain. Each variable is 
associated with a set of possible values. A configuration or a {\bf tuple $\mathbf{u'}$ of} $\mathbf{U}' \subseteq \mathbf{U}$ 
is an assignment of values to every variable in $\mathbf{U}'$. 
\end{definition}
\begin{definition}
A {\bf probabilistic domain 
model (PDM)} $P$ over $\mathbf{U}$ determines the probability $P(\mathbf{u'})$ of every tuple $\mathbf{u'}$ 
of $\mathbf{U}'$ for each $\mathbf{U}' \subseteq \mathbf{U}$.
\end{definition}
\begin{definition}
For three disjoint subsets $\mathbf{X}$, $\mathbf{Y}$ and $\mathbf{Z} \subseteq \mathbf{U}$, 
$\mathbf{X}$ and $\mathbf{Y}$ are said to be {\bf conditionally independent given 
$\mathbf{Z}$} under $P$, noted $I(\mathbf{X},\mathbf{Z},\mathbf{Y})_P$ or simply 
$I(\mathbf{X},\mathbf{Z},\mathbf{Y})$ from now on, if (see \cite[pp 83--97]{books/mk/Pearl88})
\begin{equation}
	I(\mathbf{X},\mathbf{Z},\mathbf{Y}) \equiv P(\mathbf{x}|\mathbf{y},\mathbf{z}) = P(\mathbf{x}|\mathbf{z}) 
	\;\;\text{ whenever } P(\mathbf{y},\mathbf{z}) > 0
\end{equation}
\label{def:cond_ind}
\end{definition}
Using this notation we can express unconditional independence as $I(\mathbf{X},\emptyset,\mathbf{Y})$, i.e.,
\begin{equation*}
	I(\mathbf{X},\emptyset,\mathbf{Y}) \equiv P(\mathbf{x}|\mathbf{y}) = P(\mathbf{x}) 
	\;\;{\text{ whenever }}P(\mathbf{y}) > 0
\end{equation*}
Note that $I(\mathbf{X},\mathbf{Z},\mathbf{Y})$ implies the conditional independence of all pairs
of variables $\alpha \in \mathbf{X}$ and $\beta \in \mathbf{Y}$, but the converse is not necessarily true.
\begin{definition}
A {\bf Markov Blanket $\mathbf{BL}_I(\alpha)$} of an element $\alpha \in \mathbf{U}$ is any subset
$\mathbf{S} \subset \mathbf{U}$ for which (see \cite{books/mk/Pearl88})
\begin{equation}
	I(\alpha,\mathbf{S},\mathbf{U}-\mathbf{S}-\alpha) \text{ and } \alpha \notin \mathbf{S}.
\end{equation}
\label{def:markov_blanket}
\end{definition}
An intuitive interpretation of Def. \ref{def:cond_ind} would be: Once $\mathbf{Z}$ is given,
the probability of $\mathbf{X}$ will not be affected by the discovery of $\mathbf{Y}$. Or $\mathbf{Y}$
is irrelevant to $\mathbf{X}$ once we know $\mathbf{Z}$. Note that the Markov blanket condition in
Def. \ref{def:markov_blanket} is stronger than conditional independence. It is saying that not only 
that knowing $\alpha$ is irrelevant to the class, but also to the rest of the features, so $\mathbf{S}$ 
has all the information that $\alpha$ has about $C$ and all the information $\alpha$ has about
$\mathbf{U}-\mathbf{S}-\alpha$. This takes us to our definition of redundancy:
\begin{definition}
Given a set of features $\mathbf{F}$ and a class feature $C$, a {\bf redundant feature} 
$\alpha \in \mathbf{F}$ is a feature for which exists a Markov blanket  $\mathbf{S}=\mathbf{BL}_I(\alpha)$ within 
$\{\mathbf{F},C\}$ such that $\mathbf{S} \subset \mathbf{F}$.
\end{definition}

An interesting property of Markov blankets is that if we removed a feature $\alpha$ such that 
existed $\mathbf{BL}_I(\alpha) \subset \mathbf{U}$ and now we are eliminating another feature
$\beta$ such that exists $\mathbf{BL}_I(\beta) \subset \mathbf{U}-\alpha$ then we can prove that also
exists $\mathbf{BL}_I(\alpha) \subset \mathbf{U}-\beta$, we can see the proof in \cite{DBLP:conf/icml/KollerS96}. 
That is, a redundant feature remains redundant when other redundant features are removed. So if we proceed
to remove features using this criterion, we will never have to reconsider our decisions. 

%A PDM $P$ satisfies the following axioms:
%\begin{itemize}
%\item Symmetry:
%\begin{equation*}
%  I({\mathbf{X}},{\mathbf{Z}},{\mathbf{Y}}) \Leftrightarrow I({\mathbf{Y}},{\mathbf{Z}},{\mathbf{X}}) 
%\end{equation*}
%\item Decomposition:
%\begin{equation*}  
%  I({\mathbf{X}},{\mathbf{Z}},{\mathbf{Y}} \cup {\mathbf{W}}) \Rightarrow I({\mathbf{X}},{\mathbf{Z}},{\mathbf{Y}}) \wedge %I({\mathbf{X}},{\mathbf{Z}},{\mathbf{W}}) 
%\end{equation*}
%\item Weak union:
%\begin{equation*}
%  I({\mathbf{X}},{\mathbf{Z}},{\mathbf{Y}} \cup {\mathbf{W}}) \Rightarrow I({\mathbf{X}},{\mathbf{Z}} \cup {\mathbf{W}},{\mathbf{Y}})
%\end{equation*}
%\item Contraction:
%\begin{equation*}
%  I({\mathbf{X}},{\mathbf{Z}},{\mathbf{Y}}) \wedge I({\mathbf{X}},{\mathbf{Z}} \cup {\mathbf{Y}},{\mathbf{W}}) \Rightarrow %I({\mathbf{X}},{\mathbf{Z}},{\mathbf{Y}} \cup {\mathbf{W}}) 
%\end{equation*}
%\item Intersection (holds when P is strictly positive, i.e. $P(\mathbf{u'}) > 0$, for each tuple 
%$\mathbf{u'}$ of each $\mathbf{U'} \subseteq \mathbf{U}$):
%\begin{equation*}
%  I({\mathbf{X}},{\mathbf{Z}} \cup {\mathbf{W}},{\mathbf{Y}}) \wedge I({\mathbf{X}},{\mathbf{Z}} \cup {\mathbf{Y}},{\mathbf{W}}) \Rightarrow %I({\mathbf{X}},{\mathbf{Z}},{\mathbf{Y}} \cup {\mathbf{W}})  
%\end{equation*}
%\end{itemize}

Unfortunately, there we rarely find a fully redundant feature, but rather one that its information is 
nearly subsumed by other features. So we would like to know not only whether a feature is redundant or
not but its redundancy grade. We would like a function $R'$ which given an feature $\alpha \in \mathbf{U}$ and a set of 
features $\mathbf{U} \in \mathcal{U}$ gives us a degree of redundancy of this feature to the set. Ideally we would like
a function $R' : \mathbf{U} \times \mathcal{U} \rightarrow \left[0,1\right]$ than satisfies the following propositions:

\begin{equation*}
\begin{gathered}
  R'(\alpha,\mathbf{BL}_I(\alpha)) = 1 \\
  R'(\alpha, \mathbf{U}-\alpha_i ) \leq R'(\alpha,\mathbf{U}), \forall \alpha_i  \in \mathbf{U} 
 \end{gathered} 
\end{equation*}

To achieve this we should change the boolean definition of conditional independence to a some function
of $P(\mathbf{x}|\mathbf{y},\mathbf{z})$ and  $P(\mathbf{x}|\mathbf{z})$.

\begin{definition} 
If we have that: $\mathbf{U}$ is our set of features, $\alpha$ is the feature we are evaluating, and 
$\mathbf{S}$ is some subset of $\mathbf{U}$ not containing $\alpha$. We defined $\mathbf{u}$ as a 
configuration of $\mathbf{U}$. We will write $\mathbf{s_u}$, $\mathbf{s^{-1}_u}$ and $\alpha_\mathbf{u}$ 
for the configuration of $\mathbf{S}$, the configuration of $\mathbf{U}-\mathbf{S}-\alpha$ and the 
value of $\alpha$ respectively when the configuration of $\mathbf{U}$ is $\mathbf{u}$. 
Now we can define $\mathcal{U}$ as the set of all possible configurations of $\mathbf{U}$ for which 
$P(\mathbf{u}-\mathbf{s_u}-\alpha_\mathbf{u},\mathbf{s_u})>0$. 

With all that, we define {\bf Redundancy level $R'$} as:   
\begin{equation*}
R'(\alpha,\mathbf{U}) = 1-\max_{\mathbf{S} \subset \mathbf{U}-\alpha}
{\left(  \frac{\sum_{\mathbf{u} \in \mathcal{U}}
		{ \left|P(\alpha_{\mathbf{u}}|\mathbf{s_u}) - 
		P(\alpha_{\mathbf{u}}|\mathbf{s^{-1}_u},\mathbf{s_u})\right| }}{|\mathcal{U}|}  \right)} 
\end{equation*} 
\label{def:R}
\end{definition}

Note that the calculation of this redundancy level is exponential in the number of features in our set, 
as it compares the conditional probabilities of all possible subsets of $\mathbf{U}$, so the $\max$ 
function will have to compare $|\mathcal{P}(\mathbf{U})|=2^{|\mathbf{U}|}$ terms. And for each subset
we also have an exponential cost in the number of values of the features because the sum is over each
configuration $\mathbf{u}$ of $\mathbf{U}$. 

It is clear to see that, although Eq. \ref{def:R} gives an intuitively consistent definition of redundancy
level, its computational cost might be too large for $R'$ to be directly applied in a feature weighting 
(or feature selection) algorithm. We should use an estimation of $R'$ that maximized the tradeoff between
accuracy and complexity. But in fact the aim of the definition of $R'$ was not to have an efficient algorithm to 
calculate the redundancy level of a feature. The definition had three basic (related) objectives: first 
of all to provide a suitable formal definition of redundancy in order to study the effect of feature redundancy 
in the different existing algorithms, for instance ReliefF. And second to serve as some starting point for new 
extensions to methods which performance decreases in the presence of redundant features, again Relief is an 
example. And finally, to direct the developing of new algorithms that effectively and efficiently estimate 
redundancy.
   
\subsection{Double Relief}	
When more and more irrelevant features are added to a dataset the distance calculation of Relief
degrades its performance as instances may be considered neighbors when in fact they are far from
each other if we compute its distance only with the relevant features. In such cases the algorithm
may lose its context of locality and in the end it may fail to recognize relevant features.

The $\operatorname{diff}(A_i , I_1, I_2)$ function calculates the difference between the values of the feature $A_i$
for two instances $I_1$ and $I_2$. Sum of differences over all features is used to determine the
distance between two instances in the nearest hit and miss calculation (see equation \ref{eq:relief_distance}).

As seen in the k-nearest neighbors classification algorithm (kNN) many weighting schemes 
which assign different weights to the features in the calculation of the distance 
between instances (see equation \ref{weighted_distance}). 
\begin{equation}
\delta' (I_1 ,I_2 ) = \sum\limits_{i = 1}^a {w(A_i)\operatorname{diff}(A_i , I_1 ,I_2)}
\label{weighted_distance}
\end{equation}

In the same way that in \cite{DBLP:journals/air/WettschereckAM97} Relief's estimates of features' 
quality have been used successfully as weights for the distance calculation of kNN  we could use their estimation
in the previous iteration to compute the distance between instances while searching the nearest hits and misses.
We will refer to this version of ReliefF as double ReliefF or in short dReliefF.
The problem using the weights estimates could be that in early iterations these estimations could be too biased 
to the first instances and could be far from the optimal weights. So, for small $t$, $W[A_i]$ is very different 
from $W[A_i]_t$.

What we want is to begin the distance calculation without using the weight estimates and then, 
as Relief's weight estimates become more accurate (because more instances have been taken into account), increase
the importance of these weights in the distance calculation. Lets have a distance calculation like the one in 
equation \ref{progressively_weighted_distance}.
\begin{equation}
\delta (I_1 ,I_2) = \sum\limits_{i = 1}^a {f(W(A_i)_t,t)\operatorname{diff}(A_i , I_1 ,I_2)}
\label{progressively_weighted_distance}
\end{equation}

We would like a function $f:\mathbb{R} \times (0,\infty) \rightarrow \mathbb{R} $ such that:
  \begin{itemize}
  \item $f(w,t)$ is increasing with respect to $t$
  \item is continuous
  \item $f(w,0) = 1$
  \item $f(w, \infty) = w$
  \end{itemize}

\begin{figure}[htbp] 
   \begin {center}
   \includegraphics[scale=0.8]{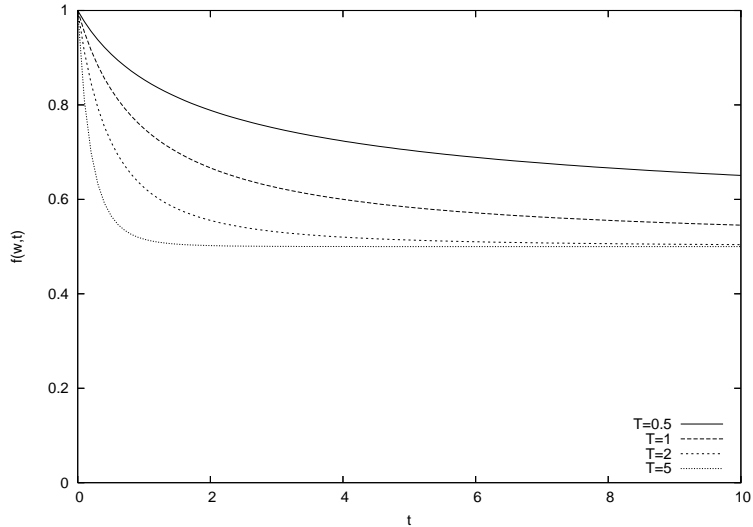}
   \end {center}
   \caption {Plot of function $f$ for 10 instances with $w=0.5$}
   \label{plotf_T}
\end{figure}

\begin{figure}[htbp] 
   \begin {center}
   \includegraphics[scale=0.8]{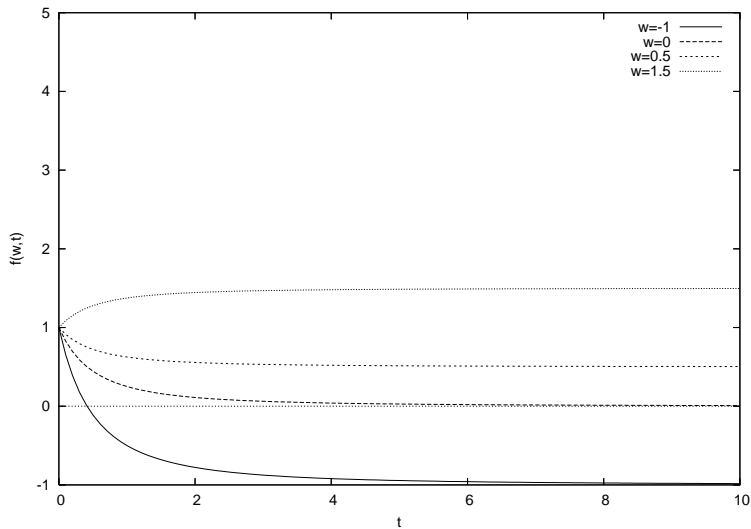}
   \end {center}
   \caption {Plot of function $f$ for 10 instances with $T=2$}
   \label{plotf_w}
\end{figure}

One such function could be the one in equation \ref{f}. And we will refer to the version of ReliefF using
this distance equation as progressively weighted double relief or in short pdReliefF.
\begin{equation}
f(w,t) = \frac{{-w + 1}}{{t^T}} + w 
\label{f}
\end{equation}
Where $T$ is a control parameter that determines the steepness of the curve described by $f$ 
(see figure \ref{plotf_T}). 
Another desirable property for our function would be that it always gives the same results regardless
of the number of iterations. In other words, if $m$ is the total number of iterations, we would
like $f(w,m)$ to be the same value whatever the value of $m$. To achieve that we must vary the value 
of $T$ according to the total number of iterations so as to decrement the steepness of the function
as the number of total iterations increases. The value of $T$ for $f(w,m)$ to be the same is
$T = 2/\log (m)$. In figure \ref{plotf_w} we can see how $f$ varies the influence of different weights
(even a non realistic one that is greater than 1) as iterations go on. We can see that with this value
for $T$ the function converges in the first few iterations and then it stabilizes its value near $w$.
For problems with many iterations a softer function may be tried if values converge prematurely.

\section{Empirical results}\label{sec:empirical}
To begin with the empirical results we have to define a measure of success for the weights estimations.
First of all we need to have a success criterion. 
For problems where we know which of the features are important (e.g., artificial datasets) some
we can use this knowledge to evaluate estimates. In \cite{DBLP:conf/ecml/Kononenko94} 
\textit{separability} and \textit{usability}, two more indicators may be useful in case of negative 
separability - \textit{minimality} and \textit{completeness} - which can help in determining the quality
of the given solution. See more precise definitions below.
\begin{description}
\item[separability] Shows the ability of the weight estimates to distinguish between 
					important and unimportant features. Positive separability ($s>0$) means that important 
					features are correctly separated from unimportant ones. 
					
					$s = W_{I_{\text{worst}}}-W_{R_{\text{best}}} \in [-2,2]$
\item[usability]	Shows the ability of the weight estimates to distinguish on of the
					important feature from the unimportant ones.  Positive usability ($u>0$) means that 
					almost one of the important features is correctly separated from unimportant ones.
					
					$u = W_{I_{\text{best}}}-W_{R_{\text{best}}} \in [-2,2]$
\item[minimality]	Shows the ratio of important features in the minimum set of features that contains
					all the important features if we select features in decreasing weight order. Note that 
					$s>0 \Rightarrow m=1$. 
					
					$m = |\mathcal{I}|/|\mathcal{M}| \in (0,1] \text{ where } \mathcal{M}=\{F|W_{F} \geq W_{I_{\text{worst}}}\}$
\item[completeness] Shows the ratio of important features that we would take if selecting features in decreasing
					weight order we stopped before selecting the first unimportant feature. Note that again
					$s>0 \Rightarrow m=1$.
					
					$c = |\mathcal{C}|/|\mathcal{I}| \in (0,1] \text{ where } \mathcal{C}=\{F|W_{F} > W_{R_{\text{best}}}\}$
\end{description}

The first set of artificial problems to use is the so-called \textit{Modulo-p-I}. In these datasets we will find 
$I$ important features and $R$ random ones. All of them integers in the range [0,p). The class value $C$ is also an
integer in the same range and can be calculated for an instance $X$ having values $X_1, X_2, \hdots, X_I$ in its
important features as seen on Eq. \ref{eq:sum_mod}. We will test our criteria for various parameters of ReliefF on
two different problems (Modulo-2-2 and Modulo-4-3) incrementally adding random features.
\begin{equation}
C(X) = \left( \sum\limits_{i=1}^I{X_i} \right) \mod p
\label{eq:sum_mod}
\end{equation}
\begin{figure}[htbp]                                                                     
  \begin{center}                                                                                                                         
  \includegraphics[scale=0.8]{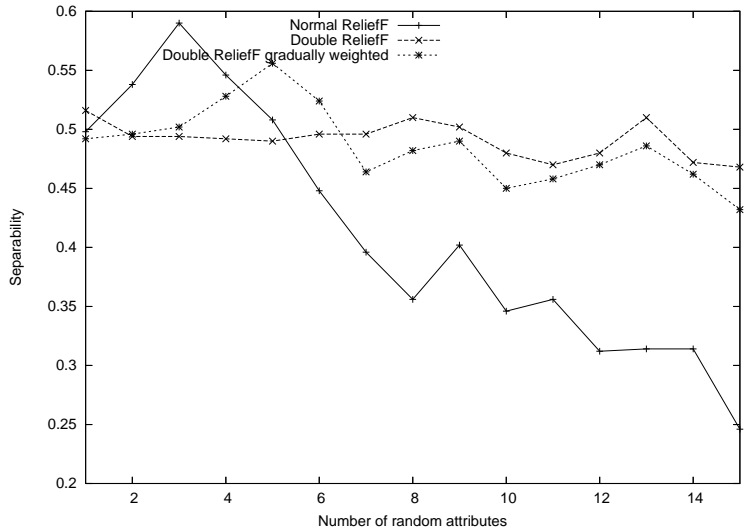}                                     
  \caption{Separability for Modulo-2-2 problem with random features}                    
  \label{fig:mod_2_2}                                                                
  \end{center}                                                                        
\end{figure}
\begin{figure}[htbp]                                                                     
  \begin{center}                                                                                                                         
  \includegraphics[scale=0.8]{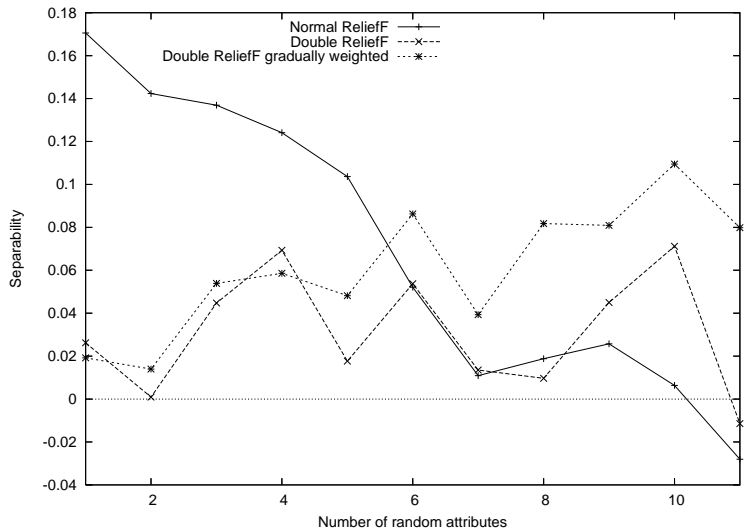}                                     
  \caption{Separability for Modulo-4-3 problem with random features}                    
  \label{fig:mod_4_3}                                                                
  \end{center}                                                                        
\end{figure}       
In Fig. \ref{fig:mod_2_2} and \ref{fig:mod_4_3} we can see the different behaviors of the three algorithms
when more and more random features are added. While ReliefF seems to gradually degrade its performance, 
dReliefF is more erratic and pdReliefF obtains the best results. This supports our theory that although it
seems a good idea to use ReliefF's own estimates as weights for its distance function, a bad start can 
make dReliefF's estimates even poor than ReliefF's. 

Another good test is CorrAl dataset introduced in \cite{DBLP:journals/ai/KohaviJ97}. This dataset is composed
of 6 features ($A_0$, $A_1$, $B_0$, $B_1$, $C$, $I$). $C$ is 75\% correlated with the class and four other features 
that can fully determine the class of the instance when used together. The class can be expressed as:
$(A_0 \wedge A_1) \vee (B_0 \wedge B_1)$. And the last one, $I$, is completely random. Results are shown in table 
\ref{taula_corral}.
\begin{table}[htbp]
\centerline{
\begin{tabular}{|c|ccc|}
\hline
Feature    &   ReliefF  & dReliefF & pdReliefF    \\
\hline \hline
$B_0$	&	0.259	&	0.272	&	0.272	\\
$B_1$	&	0.197	&	0.273	&	0.273	\\
$A_0$	&	0.194	&	0.277	&	0.278	\\
$A_1$	&	0.128	&	0.277	&	0.278	\\
$C$	&	0.281	&	0.042	&	0.044	\\
$I$	&	-0.141	&	-0.222	&	-0.222	\\
\hline \hline							
separability	&	-0.153	&	0.230	&	0.228	\\
usability	&	0.422	&	0.047	&	0.050	\\
\hline
\end{tabular}}
\caption{Weights and separability for CorrAl dataset. (With 5 nearest neighbors).}
\label{taula_corral}
\end{table}
So for CorrAl dataset the three algorithms correctly identify the irrelevant feature and rank it last, but
the normal version of ReliefF give a larger weight to the correlated feature than it should be given.
The \textit{double} versions of the algorithm in the other hand correctly identify the four features that
completely determine the class and give them larger weights, followed by the correlated one and leaving the
random one last. We can see that the behavior of the two \textit{double} versions is very similar, although
the progressive weighted estimation is a little more usable, it's a little less separable.

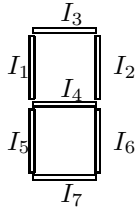
\begin{figure}[htbp] 
   \begin {center}
   \setlength{\unitlength}{2500sp}%
\begin{picture}(1203,1935)(3751,-2311)
\put(4276,-1261){\makebox(0,0)[lb]{\smash{{{$I_4$}%
}}}}
\thinlines
{\put(4014,-1355){\framebox(615,38){}}
}%
{\put(4629,-1278){\framebox(38,615){}}
}%
{\put(3976,-2009){\framebox(38,616){}}
}%
{\put(4629,-2009){\framebox(38,616){}}
}%
{\put(4014,-2086){\framebox(615,38){}}
}%
{\put(4014,-624){\framebox(615,38){}}
}%
\put(4276,-511){\makebox(0,0)[lb]{\smash{{{$I_3$}%
}}}}
\put(4276,-2311){\makebox(0,0)[lb]{\smash{{{$I_7$}%
}}}}
\put(4801,-1036){\makebox(0,0)[lb]{\smash{{{$I_2$}%
}}}}
\put(4801,-1786){\makebox(0,0)[lb]{\smash{{{$I_6$}%
}}}}
\put(3751,-1036){\makebox(0,0)[lb]{\smash{{{$I_1$}%
}}}}
\put(3751,-1786){\makebox(0,0)[lb]{\smash{{{$I_5$}%
}}}}
{\put(3976,-1278){\framebox(38,615){}}
}%
\end{picture}%
   \end {center}
   \caption {A LED display indicating the meaning of the features}
   \label{fig:led7}
\end{figure}

The next dataset (led24) is one of the LED display domain datasets from \cite{Hettich+Blake+Merz:1998}. In
fact it is an extension of the led7 dataset. The led7 dataset consists of 7 boolean valued features 
($I_1$, $\hdots$ , $I_7$) each of them representing one of the light-emitting diodes contained on a LED 
display. They indicate whether the corresponding segment is on or off (see Fig. \ref{fig:led7}). And the 
class feature has range [0,9] and coincides with the digit represented by the display. This dataset has
another added difficulty as it has a 10\% of noise in its features, i.e., each instance's feature has
a 10\% chance of having its value negated. This is a quite difficult problem for classifiers and the 
version with 17 unimportant features is especially difficult, e.g. a nearest neighbor classification 
algorithm falls from a 71\% of classification success with the 7 feature version to a poor 41\% with the 
other one. So it would be desirable for ReliefF to separate the important features from the rest.
\begin{table}[htbp]
\centerline{
\begin{tabular}{|c|cc|}
\hline
Algorithm	&	$s$	&	$u$ \\
\hline \hline				
ReliefF	&	0.131	&	0.340 \\
dReliefF	&	0.084	&	0.234 \\
pdRelefF	&	0.104	&	0.278 \\
\hline
\end{tabular}}
\caption{Separability and usability for led24 dataset.}
\label{taula_led24}
\end{table}
Table \ref{taula_led24} shows separability and usability for this dataset. There it can be seen that
the behavior for the three algorithms is extremely similar for this domain. All of them are able to
separate the seven important features from the rest and even the values for $s$ and $u$ are almost 
the same for the three algorithms.

Finally, the last artificial datasets to be tested are Monks datasets. They are interesting because 
even though they do not consist of lots of features, they are well known datasets, have interesting
feature interactions and can serve us to compare the algorithms order of each feature with its intended
ordering. There are three Monks datasets but we will only use Monk-1 and Monk-3 because Monk-2
does not contain unimportant features. They consist of six numerical features $A_1, \hdots, A_6$ 
with ranges varying from [1,2] to [1,4] and a boolean class value. For Monk-1 the class $C_{M1}$ can 
be calculated as $C_{M1}=(A_1=A_2) \vee (A_5=1)$ and the class $C_{M3}$ for Monk-3 as 
$C_{M3}=(A_5 = 3 \wedge A_4 = 1) \vee (A_5 \neq 4 \wedge A_2 \neq 3)$. So for the first problem, $A_3$,
$A_4$ and $A_6$ are unimportant and among the other three, $A_1$ and $A_2$ would help us better
determine the class value than $A_5$ as only one of the four possible values of $A_5$ is important. 
For Monk-3 the important features will only be $A_5$, $A_4$ and $A_2$ and the rest do not influence 
the instance's class. Among these three features, $A_5$ and $A_2$ should be preferred over $A_4$ as 
using only the second term of the disjunct we can achieve a 97\% performance.
It is important to say that Monk-3 has a 5\% of additional noise (misclassifications).
\begin{table}[htbp]
\centerline{
\begin{tabular}{|c|ccc|}
\hline
Algorithm		$s$		$u$		Feature ordering \\
\hline \hline						
ReliefF	&	0.26	&	0.38	&	$A_1$,$A_2$,$A_5$,$A_3$,$A_6$,$A_4$ \\
dReliefF	&	0.42	&	0.44	&	$A_5$,$A_1$,$A_2$,$A_3$,$A_6$,$A_4$ \\
pdReliefF	&	0.41	&	0.43	&	$A_1$,$A_5$,$A_2$,$A_3$,$A_6$,$A_4$ \\
\hline
\end{tabular}}
\caption{Separability, usability and feature ordering for Monk-1 dataset.}
\label{taula_monks1}
\end{table}
\begin{table}[htbp]
\centerline{
\begin{tabular}{|c|ccc|}
\hline
Algorithm		$s$		$u$		Feature ordering	\\
\hline \hline							
ReliefF	&	0.05	&	0.43	&	$A_5$,$A_2$,$A_4$,$A_3$,$A_1$,$A_6$	\\
dReliefF	&	0.08	&	0.29	&	$A_2$,$A_5$,$A_4$,$A_3$,$A_1$,$A_6$	\\
pdReliefF	&	0.05	&	0.31	&	$A_2$,$A_5$,$A_4$,$A_3$,$A_1$,$A_6$	\\
\hline
\end{tabular}}
\caption{Separability, usability and feature ordering for Monk-3 dataset.}
\label{taula_monks3}
\end{table}
Table \ref{taula_monks1} shows the results for the three variants of ReliefF when applied to the Monk-1
dataset. We observe that although the three algorithms correctly separate the important features from the
unimportant ones, only ReliefF gives the expected ordering for the important features. The same results 
for Monk-3 dataset are shown in table \ref{taula_monks3}. For this dataset we can see how separability, 
even though positive, is very small for the three algorithms. In addition, all of them rank $A_4$ as the 
lowest of the important features which agrees with what we thought they should do. Here the \textit{double}
versions of the algorithms seem to help discriminating the important from the unimportant features, in 
the two cases they improve separability although the important feature order is worse. The \textit{double}
versions seem to increase the weight difference between important and unimportant features but decrease the 
weight difference of features in the same group. 

The second group of experiments is with some well known datasets from UCI \cite{Hettich+Blake+Merz:1998}.
These are datasets of real data, so we don't know which of the features may be important and which +
may be not. For this reason we will not be able to compute the above criteria for these datasets. So
to evaluate the quality of the algorithms' estimates we will use the performance obtained with a
classifier. We will make various tests with the classifier. We will first of all try a classification 
with the feature with the greatest weight, then will use the two most weighted variables, end so on until
all variables are used. When all tests are completed we will compare the performance of the classifier
when using all features with the performance when using the best subset found using Relief's estimates.
We will use the 1NN classifier because of its simplicity and sensibility to a bad choice of features.

\begin{figure}[htbp] 
   \begin {center}
   \includegraphics[scale=0.8]{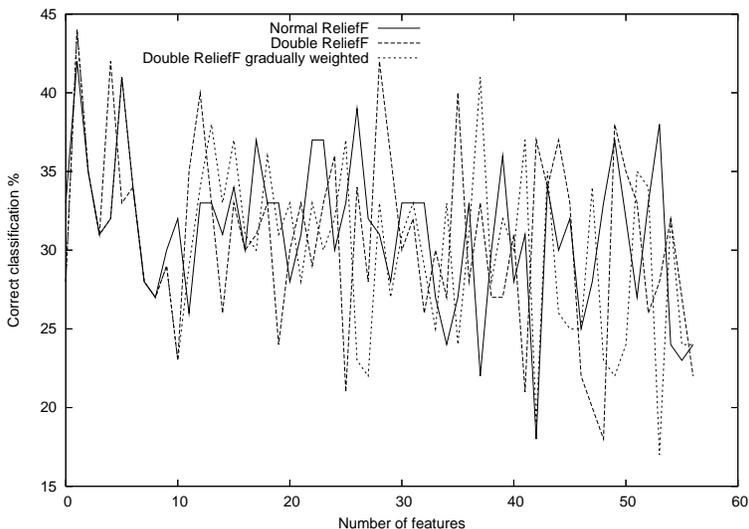}
   \end {center}
   \caption {Classification success \% with for the promoter gene problem}
   \label{fig:promoters}
\end{figure}

The first chosen dataset is the E. coli promoter gene sequences. This dataset contains a set of 57 
nominal variables representing a DNA sequence of nucleotides. A promoter is a DNA sequence that enables 
a gene to be transcribed. The promoter is recognized by RNA polymerase, which then initiates transcription.
For the RNA polymerase to make contact, the DNA sequence must have a valid conformation so that the two
pieces of the contact region spatially align. But shape of the DNA molecule is a very complex function of
the nucleotide sequence due to the so complex interactions between them, so strong interactions among
features are expected. In Fig. \ref{fig:promoters} we can see the results of applying feature selection
in the way described above for the 1-NN classifier.
As can be seen the results for the classification task are in general not very good, but we can see that for
all the three versions of the algorithm the maximum performance is achieved when the number of used features 
is 2, much less than the initial 57 features. The three versions of the algorithm agree in the first two 
features to be add (15 and 16), although the ordering in the case of normal Relief is inverted it selects 16 
first and then 15.

Another problem that can serve us to determine whether the weighted distance calculation makes sense is the 
lung cancer dataset also from UCI. It consists of data from 32 patients suffering three different types of
pathological lung cancers. The objective is to distinguish among the three types of cancer given a set of 56
nominal features with ranges [0,3]. Authors of the dataset gave no information on the meaning of individual
features. But probably data may be from different types of tests performed on patients and as there are 
many features one can venture the hypothesis that many of them may be standard tests that can not help in 
determining the patient's type of disease. So these unimportant features may affect the way that ReliefF 
chooses the nearest neighbors. Moreover, it is especially important to reduce the number of features in this
problem because the number of instances is very low compared to it so classifiers may be fooled by unimportant
features.
\begin{figure}[htbp] 
   \begin {center}
   \includegraphics[scale=0.8]{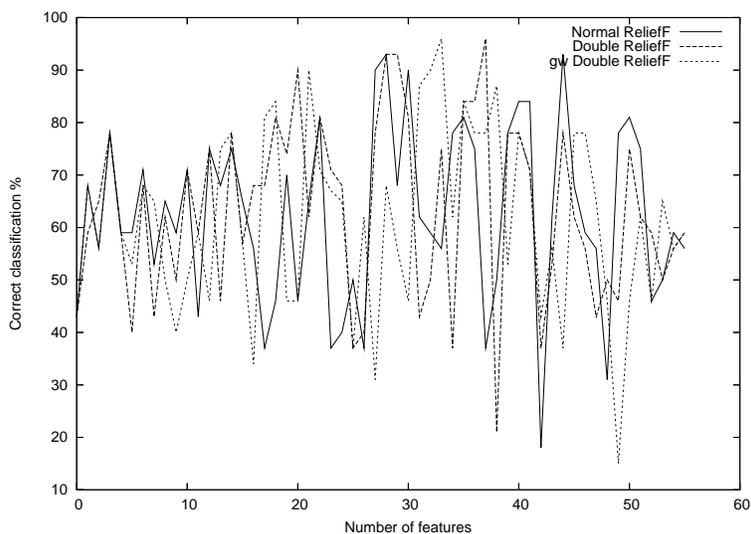}
   \end {center}
   \caption {Classification success \% for the lung cancer problem}
   \label{fig:lung}
\end{figure}
\begin{table}[htbp]
\centerline{
\begin{tabular}{|c|cc|}
\hline
Algorithm	&	Max. 1-NN Performance	&	\# of features \\
\hline \hline				
ReliefF	&	93.75	&	28 \\
dReliefF	&	96.88	&	37 \\
pdRelefF	&	96.88	&	33 \\
\hline
\end{tabular}}
\caption{Best results obtained with 1-NN classifier for the lung cancer problem.}
\label{taula_lung}
\end{table}
We can see in Fig. \ref{fig:lung} that in this case performance of the 1-NN classifier is significantly 
improved when we apply feature selection for this problem. While a classification using all of the
features gives us a correct classification percent of 43.75, the bests results obtained with a subset of
the features is above 90\% with all of the versions of ReliefF. Although the same performance is achieved
with the best subset given by dReliefF and the one given by pdReliefF, the results of the latter are
better as the same performance is achieved with 4 less features. 

\section{Conclusions and future work}\label{sec:conclusions}
In our experiments we have seen how the {\it double} versions of the algorithm helped in the correct feature
weighting of some problems while in other cases performance is not improved and even it is diminished.
An interesting property of these new versions of the algorithm is than they seem to help in problems where
many irrelevant features exist, which was the initial objective. The performance of the algorithm improved 
in the modulo-p-I problems as more and more random features were added. We saw in the experiments that although
ReliefF's performance with few attributes was better, as the number of random features increased it began
to decrease and for a relatively small number of random features dReliefF and pdReliefF overperformed the 
original algorithm. Furthermore the performance of the latter methods did not vary with the addition of random
features. In contrast, the results obtained with the LED dataset were not that encouraging. Although the dataset
had more than twice random features than relevant ones the results for the three algorithms were very similar. 
This might be because the separability for this problem was so low (though positive) due to the difficulty of the 
problem (even without random features) for the presence of noise. The try with datasets having fewer irrelevant
features, i.e. the Monks problems, gave very similar results for all the versions. This has a logical explanation: 
the behavior of the {\it double} version if all the attributes are relevant is not very different from the original 
one. 

The experiments with real data from the UCI Machine Learning Repository \cite{PRNN:Hunt+Marin+Stone:1966} that 
consisted in running a 1-NN classifier using successive subsets of features proposed by the three versions of ReliefF
showed interesting results. The success evaluation criteria was the percentage of instances classified correctly
using 5-fold crossvalidation. Two datasets were chosen for this experiments because of their large number
of features and the intuition that they might contain large number of irrelevant features. In both of them the
{\it double} versions of the algorithm chose a subset of features that helped the 1-NN best in classifying the
instances. In the case of the DNA promoters dataset the performance increase was not significant but this may be
due to the fact that 1-NN do not seem to be capable of solving this problem as it gave poor results in all cases.
On the other hand, for the lung cancer dataset, we obtained significantly better classifying performance with the 
subset from the {\it double} versions and, in addition, the subset found by the pdReliefF had less variables than 
the one found by dReliefF.

We experienced almost no difference between the two {\it double} versions. This may be because of the progressive
weighting function used. The function attenuates the weight estimates influence at first iterations but rapidly
increases their influence and after the first few iterations the algorithm behaves exactly as dReliefF. So as
a future work some other \textit{softer} functions may be tested. 

Another clear line of future work is the formal study of the influence that having redundant features has to 
ReliefF, dReliefF and pdReliefF. Robnik-\v{S}ikonja and Kononenko started this study in 
\cite{DBLP:journals/ml/Robnik-SikonjaK03} where they proved that the addition of successive copies of one 
feature divided the initial weight ReliefF assigned to the feature among all the copies. And they received 
the same weight. But still some crucial questions have to be answered: Do equal weights for two features
mean that features are redundant to each other? Does an equal sequence of weight actualizations for two 
features mean that they are redundant to each other? How can Relief be extended to diminish or eliminate the 
negative effect of redundant features? Does ReliefF compute some kind of approximation to $R'$?

\bibliography{relief}

\end{document}